\documentclass[runningheads]{llncs}
\usepackage{graphicx}
\usepackage{booktabs}
\usepackage{pifont}
\usepackage[colorlinks]{hyperref}
\usepackage[dvipsnames]{xcolor}
\hypersetup{
    colorlinks = false,
    linkbordercolor = {blue}
}
\usepackage{listings}
\usepackage{courier}

\definecolor{codegreen}{rgb}{0,0.6,0}
\definecolor{codegray}{rgb}{0.5,0.5,0.5}
\definecolor{codepurple}{rgb}{0.58,0,0.82}
\definecolor{backcolour}{rgb}{0.95,0.95,0.92}
\definecolor{superlightgray}{RGB}{242,242,242}

\lstdefinestyle{mystyle}{
    backgroundcolor=\color{backcolour},
    commentstyle=\color{codegreen},
    keywordstyle=\color{magenta},
    numberstyle=\tiny\color{codegray},
    stringstyle=\color{codepurple},
    basicstyle=\ttfamily\footnotesize,
    breakatwhitespace=false,
    breaklines=true,
    captionpos=b,
    keepspaces=true,
    numbers=left,
    numbersep=5pt,
    showspaces=false,
    showstringspaces=false,
    showtabs=false,
    tabsize=2
}
\newcommand{\cmark}{{\color{ForestGreen}\ding{51}}}%
\newcommand{\xmark}{{\color{Maroon}\ding{55}}}%

\def\rot{\rotatebox}

\begin{document}

\title{ShabbyPages: A Reproducible Document Denoising and Binarization Dataset}
\author{Alexander Groleau\inst{1} \and
  Kok Wei Chee\inst{1} \and
  Stefan Larson\inst{2} \and
  Samay Maini\inst{1} \and
  Jonathan Boarman\inst{1}}

\authorrunning{A. Groleau et al.}

\institute{Sparkfish LLC (\email{augraphy@sparkfish.com})
  \and
  Vanderbilt University}

\titlerunning{ShabbyPages}

\maketitle

\begin{abstract}
Document denoising and binarization are fundamental problems in the document processing space, but current datasets are often too small and lack sufficient complexity to effectively train and benchmark modern data-driven machine learning models.
To fill this gap, we introduce \emph{ShabbyPages}, a new document image dataset designed for training and benchmarking document denoisers and binarizers.
\emph{ShabbyPages} contains over 6,000 clean ``born digital" images with synthetically-noised counterparts (``shabby pages") that were augmented using the \emph{Augraphy} document augmentation tool to appear as if they have been printed and faxed, photocopied, or otherwise altered through physical processes.
In this paper, we discuss the creation process of \emph{ShabbyPages} and demonstrate the utility of \emph{ShabbyPages} by training convolutional denoisers which remove real noise features with a high degree of human-perceptible fidelity, establishing baseline performance for a new \emph{ShabbyPages} benchmark.
\end{abstract}

\section{Introduction}
Denoising is a fundamental concern for many document processing workflows, wherein unwanted artifacts introduced to a document image via noisy processes like scanning are removed.
The effectiveness of the denoising stage of document processing pipelines has implications for downstream tasks like optical character recognition (OCR) and layout parsing \cite{Mustafa_2018-wan,blind-denoising-iccv-2021,patch-based-document-denoising,character-recognition-systems,ogorman-document-image-analysis,Rotman2022-hh}.
Recent work in supervised machine learning has yielded promising results at the denoising task \cite{tensmeyer-cnn-binarization-2017,blind-denoising-iccv-2021}, increasing the importance of access to large volumes of high-quality training and evaluation data.

Prior work has introduced denoising and binarization datasets, but these are often small in scale, limited in the types of noise features present, not diverse enough to train robust models, and impossible to recreate.
\begin{itemize}
\item \emph{DIBCO} datasets, which are widely recognized as critical data for document denoising and binarization, are far too small for large-scale training since each \emph{DIBCO} dataset ranges between 10-20 images total.
\item \emph{NoisyOffice} \cite{ref_NoisyOfficeDatabase} is a denoising dataset that is also quite small, having \emph{72} training samples, and contains only a few types of noise — primarily artificial wrinkles and stains, with little variation.
\item \emph{DDI-100} \cite{ddi-100-2019} is larger corpus in scale than \emph{NoisyOffice} and the \emph{DIBCO} datasets, but is again limited in the degree of feature diversity \cite{detection-masking-2022}.
\end{itemize}

In this paper, we introduce a new dataset designed primarily for the document denoising task, which addresses each of the concerns listed above.
This new dataset, \emph{ShabbyPages}, is large-scale, having \emph{6202} synthetic training images and \emph{200} real testing images.
The synthetic images were generated with the \emph{Augraphy}\footnote{\url{https://github.com/sparkfish/augraphy}} data augmentation tool, which is tailored specifically to introducing realistic noise to document images.
\emph{ShabbyPages} is more diverse than alternatives like \emph{NoisyOffice}, as it contains documents from various language groups and contains documents with graphical elements like tables and figures.
In the next sections, we provide motivation and context for \emph{ShabbyPages}' creation, compare it at length to other important datasets in the same space, describe the corpus construction process, and finally demonstrate its utility by benchmarking document denoising models trained on the dataset.

\begin{figure}
\centering
\includegraphics[width=0.98\columnwidth]{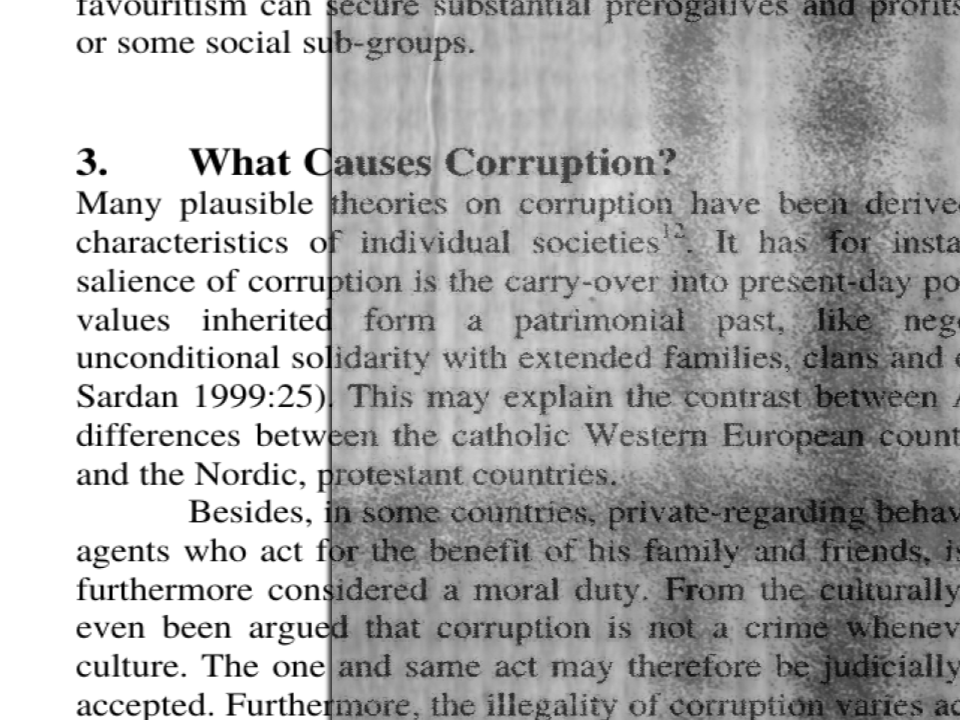}
\caption{Sample clean and corresponding dirty patch from \emph{ShabbyPages}.} \label{fig1}
\end{figure}

\section{Related Datasets}
This section discusses datasets used for the tasks of document denoising and document binarization.
These datasets are listed in Table~\ref{tab:datasets}, where we pay particular attention to dataset size, diversity of documents/texts (i.e., font sizes and styles, languages, paper styles, and graphics).
The most similar to \emph{ShabbyPages} being \emph{NoisyOffice} \cite{ref_NoisyOfficeDatabase}.
\emph{NoisyOffice} is very limited, however, consisting of two font weights, three font sizes, three font types, and four noise types, for a total of just 72 unique images.
The \emph{NoisyOffice} collection was created by permuting these font/size/base combinations, and the resulting set is feature-poor.
\emph{NoisyOffice} also lacks non-English and non-Latin characters, as well as graphical elements like images, logos, and tables.
To add noise, the creators produced four different types of background noise (folds, wrinkles, coffee stains, and footprints), and then overlayed digitally-generated text onto these.

\begin{table}[]
    \centering
    \caption{Summary of datasets for document binarization (above line) and document denoising (below line) tasks. Datasets listed below the line contain synthetic noise.}
    \label{tab:datasets}
    \begin{tabular}{lclllllll}
        \textbf{Dataset} &
        \rot{80}{\textbf{Dataset Size}}
            & \rot{80}{\textbf{Synthetic Noise}}
            & \rot{80}{\textbf{Groundtruths}}
            & \rot{80}{\textbf{Varied fonts/sizes}}
            & \rot{80}{\textbf{Varied paper styles}}
            & \rot{80}{\textbf{Multilingual}}
            & \rot{80}{\textbf{Contains graphics}}
            & \rot{80}{\textbf{Reproducible}}\\
            \midrule
         \emph{DIBCO-9} \cite{dibco-09} & 10 & \xmark & \cmark &   \cmark & \cmark & \cmark & \xmark & n/a \\
         \emph{DIBCO-10} \cite{dibco-10} & 10 & \xmark & \cmark &  \cmark & \cmark & \xmark & \xmark & n/a \\
         \emph{DIBCO-11} \cite{dibco-11} & 16 & \xmark & \cmark &  \cmark & \cmark & \cmark & \xmark & n/a\\
         \emph{DIBCO-12} \cite{dibco-12} & 14 & \xmark & \cmark &  \cmark & \cmark & \xmark & \xmark & n/a \\
         \emph{DIBCO-13} \cite{dibco-13} & 16 & \xmark & \cmark &  \cmark & \cmark & \cmark & \xmark & n/a \\
         \emph{H-DIBCO-14} \cite{dibco-14} & 10 & \xmark & \cmark & \cmark & \cmark & \cmark & \xmark & n/a \\
         \emph{H-DIBCO-16} \cite{dibco-16} & 10 & \xmark & \cmark & \cmark & \cmark & \cmark & \xmark & n/a \\
         \emph{DIBCO-17} \cite{dibco-17} & 20 & \xmark & \cmark & \cmark & \cmark & \cmark & \xmark & n/a \\
         \emph{H-DIBCO-18} \cite{dibco-18} & 10 & \xmark & \cmark  & \cmark & \cmark & \cmark & \xmark & n/a\\
         \emph{Bickley Diary} \cite{bickley-diary} & 7 & \xmark & \cmark   & \xmark & \xmark & \xmark & \xmark & n/a\\
         \emph{PHIBD} \cite{nafchi-2013-icdar} & 15 & \xmark & \cmark &   \cmark & \cmark & \xmark & \xmark & n/a\\
         \midrule
         \emph{NoisyOffice} \cite{ref_NoisyOffice} & 72  & \cmark & \cmark &  \cmark & \xmark & \xmark & \xmark & \xmark \\ 
         \emph{DocCreator} \cite{ref_DocCreator} & 6,059 & \cmark & \cmark & \cmark & \cmark & \cmark & \xmark &\cmark\\
         \emph{DDI-100} \cite{ddi-100-2019} & 100k+~ & \cmark & \cmark & \cmark & \xmark & \xmark & \cmark & \xmark\\
         \emph{ShabbyPages} (ours)~ & 6,202~ & \cmark & \cmark  & \cmark~ & \cmark~ & \cmark & \cmark & \cmark \\ 
         \bottomrule
    \end{tabular}
\end{table}

\begin{figure}
    \centering\scalebox{0.365}{
    \includegraphics{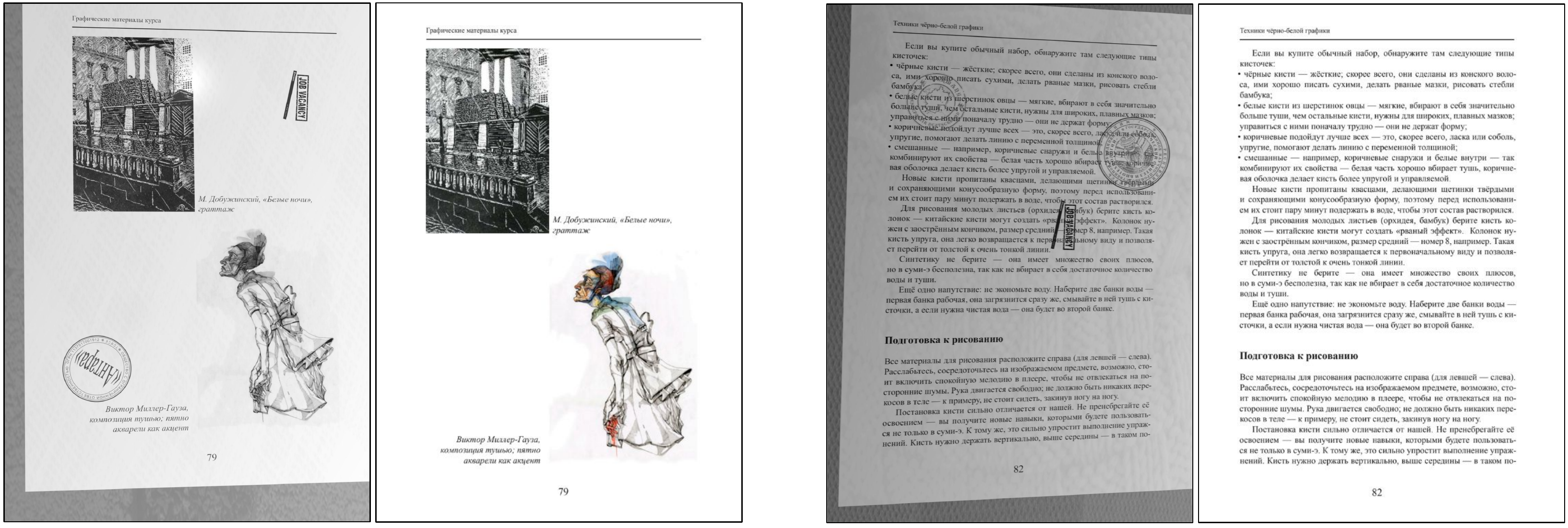}}
    \caption{Example noisy-clean pairs from DDI-100. Geometric transformations are required to map the noisy samples to their corresponding clean groundtruth images, making DDI-100 unsuitable for binarization and denoising.}
    \label{fig:ddi-100-samples}
\end{figure}

Another dataset, \emph{DDI-100} \cite{ddi-100-2019}, was constructed in a manner similar to \emph{NoisyOffice}.
\emph{DDI-100} is far larger than \emph{NoisyOffice}, but has been criticized for not containing enough diversity or noise to be an effective benchmark \cite{detection-masking-2022}.
Additionally, the pipeline code for creating the noisy versions of both \emph{NoisyOffice} and \emph{DDI-100} is not available, and thus these datasets cannot be reproduced.
Moreover, while \emph{DDI-100} has noise-free document images with noisy counterparts, the noisy images also exhibit geometric transformations --- as seen in Figure~\ref{fig:ddi-100-samples} --- making it unsuitable for benchmarking document binarizers or denoisers using standard metrics like SSIM, PSNR, or MSE, which require clean-noisy correspondences to be free of geometric transformations.

\begin{figure}
    \centering\scalebox{0.4}{
    \includegraphics{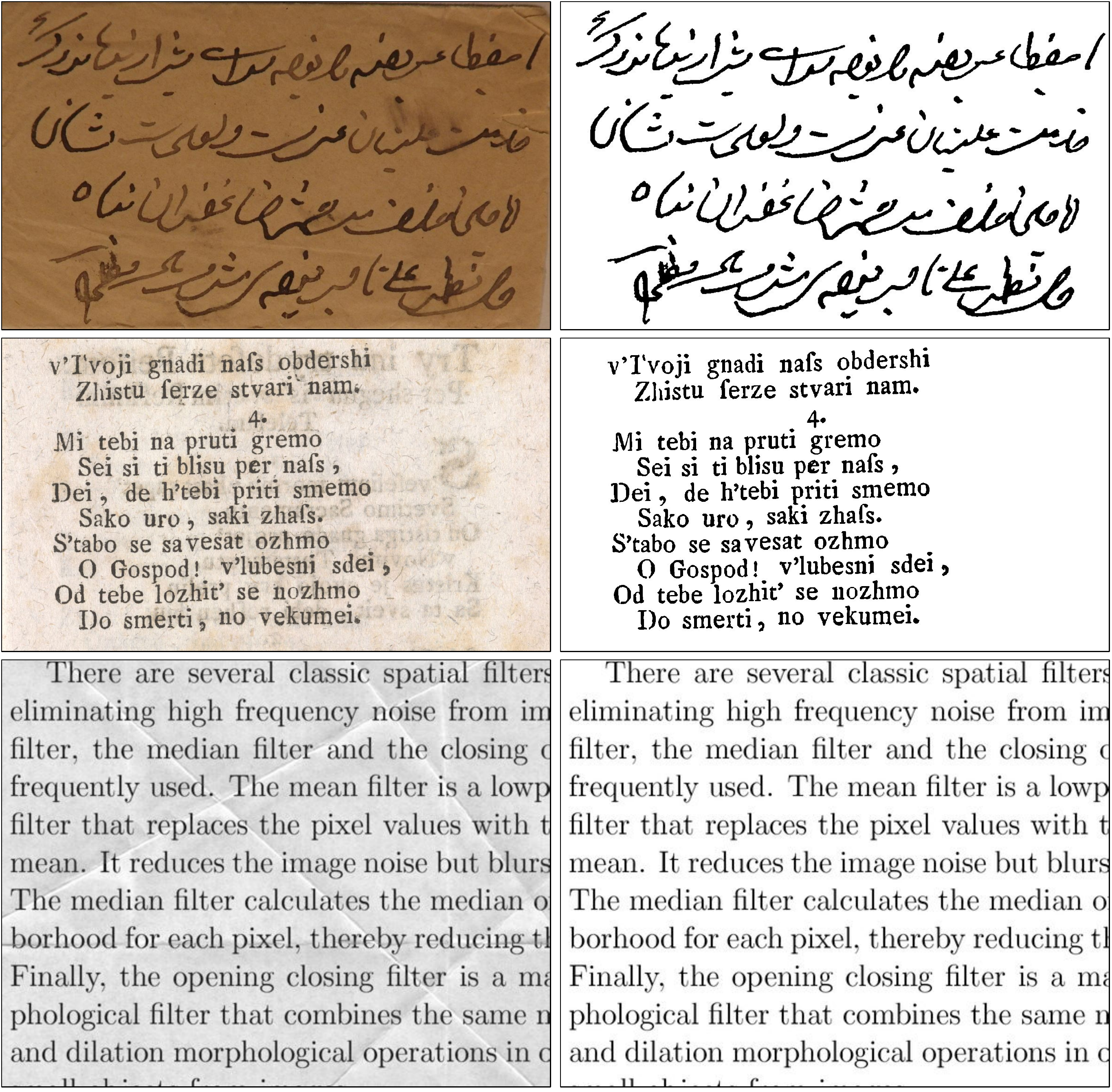}}
    \caption{Example noisy-clean pairs from PHIBD (top row), \emph{DIBCO-13} (middle row), and \emph{NoisyOffice} (bottom row).}
    \label{fig:other-dataset-examples}
\end{figure}

Other prior work includes several datasets for the related task of document image binarization, which seeks to convert a document image into a binary image, where foreground text is black and background pixels are white.
Among the most important in this class are the \emph{DIBCO} collections \cite{dibco-09,dibco-10,dibco-11,dibco-12,dibco-13,dibco-14,dibco-16,dibco-17,dibco-18}.
As summarized in Table~\ref{tab:datasets}, the \emph{DIBCO} family of datasets is very small, ranging between 10 and 20 samples each (for a total of 106 images across \emph{DIBCO-9} through \emph{DIBCO-18}).
Similar to \emph{NoisyOffice}, the \emph{DIBCO} datasets do not contain graphical elements like images, tables, charts, etc.
The Persian Heritage Image Binarization Dataset (PHIBD) \cite{phibc-2012-nafchi,nafchi-2013-icdar} is similar to the \emph{DIBCO} benchmarks, consisting of a small number of images of handwritten Persian writing.
Others, like the Bickley Diary set \cite{bickley-diary} and the 2019 Time-Quality Binarization Competition set \cite{2019-time-quality-competition} have also been used for benchmarking document binarization approaches, but these do not appear to be publicly available \cite{tensmeyer-binarization-review-2020}.
Ultimately, the sizes of these document binarization corpora severely limit the types of algorithms and models that can be used.
Examples of noisy-clean image pairs from some of these collections are displayed in Figure~\ref{fig:other-dataset-examples}.

Other popular document image datasets --- like RVL-CDIP \cite{ref_RVL-CDIP}, Tobacco-3482 \cite{tobacco-3482}, Tobacco-800 \cite{ref_Tobacco800}, and FUNSD \cite{jaume2019-funsd} --- include large amounts of real noise, but none of these have corresponding clean groundtruth document images to train or evaluate document denoisers and binarizers.

\begin{table}
    \centering
    \caption{ShabbyPages dataset statistics.}
    \label{tab:summary-statistics}
    \begin{tabular}{@{\hspace{2em}}l@{\qquad}@{\hspace{2em}}l@{\qquad}}
        \toprule
        \textbf{Statistic} & \textbf{Value} \\
        \midrule
        Num. Images & 6202 \\
        DPI & 150 \\
        Max image size & 3336x12157 \\
        Min image size & 532x532\\
    \bottomrule
    \end{tabular}
\end{table}

\section{The \emph{ShabbyPages} Dataset}
This section provides an overview of the finer details of the \emph{ShabbyPages} dataset, as well as the processes we used to create it.

\subsection{Overview of \emph{ShabbyPages}}
\emph{ShabbyPages} consists of 6,202 full-page document images.
The text in the images is varied: \emph{ShabbyPages} encompasses multiple languages (including English, Japanese, Chinese), and font styles and sizes.
Some other summary statistics for \emph{ShabbyPages} are displayed in Table~\ref{tab:summary-statistics}.
\emph{ShabbyPages} contains many more samples than the document binarization datasets surveyed in Table~\ref{tab:datasets}.
While \emph{ShabbyPages} has fewer samples than \emph{DDI-100}, we again point out that the use of \emph{DDI-100} for benchmarking document denoisers and binarizers is problematic due to the presence of geometric transformations in \emph{DDI-100}.
Our dataset contains more samples than \emph{NoisyOffice} and \emph{DocCreator},
and in contrast to \emph{NoisyOffice}, the \emph{ShabbyPages} dataset also contains graphical elements like figures, tables, logos, etc.

We can also measure the ``diversity'' of our dataset as a way to compare the richness of various features against an alternative like \emph{NoisyOffice}.
To do this, we use a diversity metric \cite{larson-etal-2019-outlier} defined as
$$
Diversity(X) = \frac{1}{|X|^2} \sum_{a\in X} \sum_{b\in X} \|f(a)-f(b)\|_2
$$
where $f(\cdot)$ is an embedding function that maps the input image 
into an \emph{n}-dimensional vector space.
Here, we use CLIP's ViT model \cite{clip}, which embeds each image into a 512-dimensional space.
The intuition behind the diversity metric is that datasets where images are highly similar will have lower diversity scores, and datasets where images share less visual similarity will have higher diversity scores.
A dataset with low diversity might not be representative enough of the real world, and low diversity may further correlate with task easiness.
Using this metric, we find that the diversity of \emph{NoisyOffice} is 0.317, while the diversity score for \emph{ShabbyPages} is 0.488.

\subsection{\emph{ShabbyPages} Construction}
This section describes the dataset generation methodology; code for all of this is available on GitHub.

\emph{ShabbyPages} was born out of a desire to address shortcomings in the popular datasets discussed in the previous section.
Reproducibility was a primary concern, so \emph{ShabbyPages} includes not only the groundtruth images, but also the original digital documents used to produce them, and the software for fully recreating the set, allowing anyone to quickly produce a new member of the \emph{Shabby} family, optionally re-exporting groundtruths at a different resolution beforehand.
The 6202 images in the \emph{ShabbyPages} release set are really just the first sample from the space of these documents; we encourage readers to build their own.
\emph{NoisyOffice} and \emph{DDI-100} are the only other datasets we examined that contain its digital provenance, but the released database materials lack a means of reproducing the simulated printing method used to create the \emph{NoisyOffice} training data; \emph{Augraphy}'s \texttt{PaperFactory} augmentation is such a tool.
The other sets are not even theoretically extensible in this way.

\begin{figure}
    \includegraphics[width=1.0\textwidth]{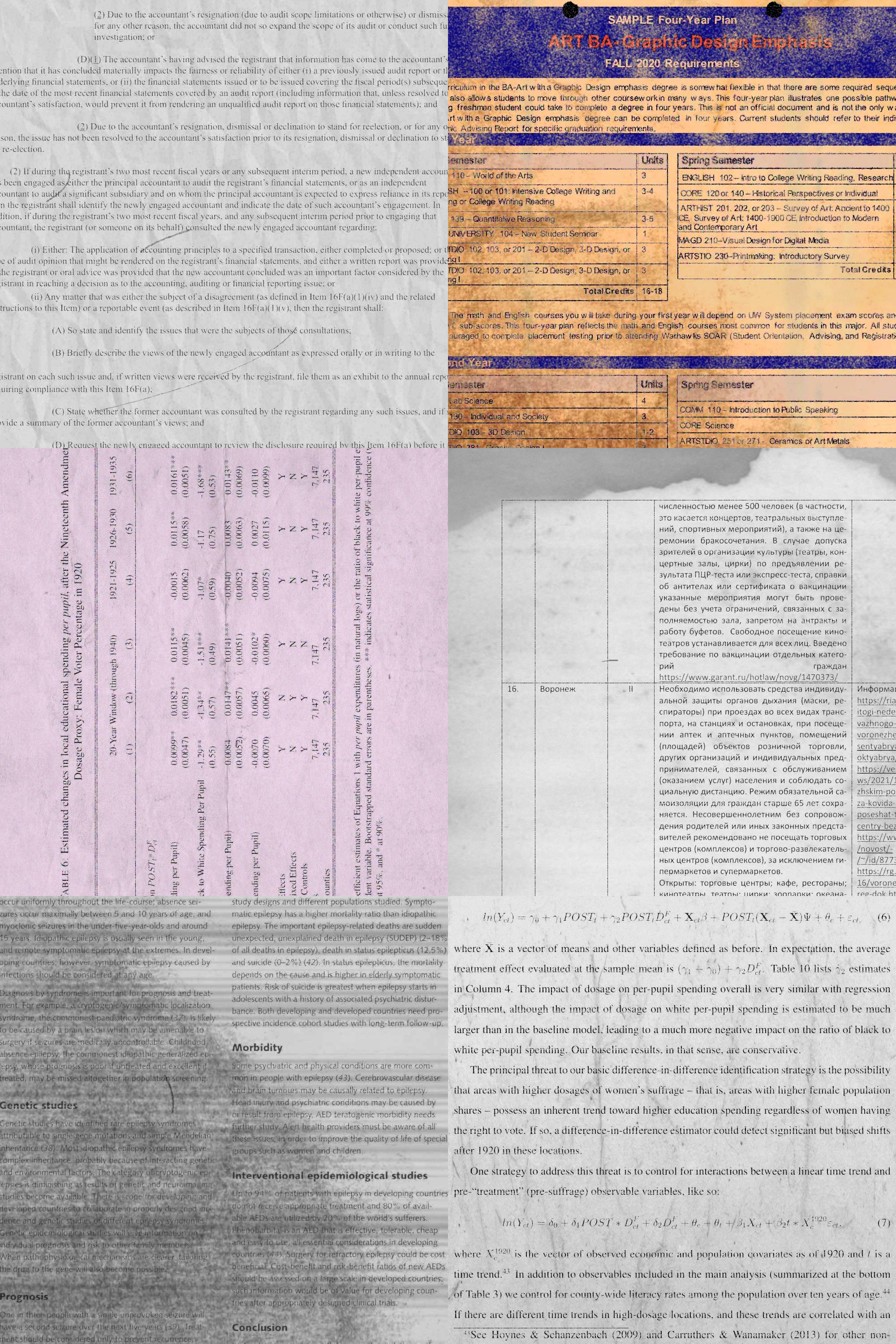}
    \caption{Sample noisy patches from \emph{ShabbyPages}.}
    \label{fig:matrix_tall1}
\end{figure}

\subsubsection{Data Gathering.}
A team of crowdworkers searched the public internet for PDFs of many different kinds, from culturally-important sources like government agencies, NGOs, and large multinational corporations.
A set of 600 unique documents were retrieved, totaling 6202 pages.
Care was taken to retrieve freely-available and attributable documents; a manifest is included in the dataset which contains an ordered table of the filenames, the languages present in the document, the download URL for the file, and an English name or acronym for the source organization or person.
A similar process was completed to collect paper textures on which to ``print'' the documents: 300 unique textures were gathered, all of which are either in the public domain or carry permissive Creative Commons licenses (\texttt{CC-0}, \texttt{CC-BY}, or \texttt{CC-BY-SA}); metadata about these textures appears in another manifest.
We also manually reviewed the gathered documents to verify that no personal identity information was present within the corpus.

\subsubsection{PDF to PNG Conversion.}
The \texttt{pdftoppm} tool from the \texttt{poppler-utils} package was used to split the PDF documents into individual pages.
Each page was converted to a PNG image at 150dpi.
Because the majority of these documents were created with the standard US Letter dimensions, this resulted in the most common image dimension being 1275 pixels wide by 1650 pixels tall.

\subsubsection{Developing the Pipeline.}
The \emph{Augraphy} library presents an easy-to-use API for constructing data augmentation pipelines for document images, which has been designed for interoperability with other augmentation tools, and within the broader data ecosystem.
While \emph{Augraphy}'s default pipeline has what we believe to be quite realistic defaults, we wanted a broader range of features from this dataset than those the default pipeline could produce.
To address this, we broke all of the parameters for every augmentation constructor out into separate variables, tweaking these and committing the new pipeline to GitHub.
We created an automated daily build in GitHub Actions to render a random selection of groundtruth images with the updated pipeline, then our team met frequently to discuss the output and make adjustments to the augmentation parameters, until we were consistently satisfied with the results.

The pipeline uses 23 of the augmentations from \emph{Augraphy}.  Augmentations such as \texttt{BookBinding} and \texttt{Folding} were excluded since these spatial transformations alter the document image geometry by shifting pixel placement from their original position.  Such transforms would render images unsuitable for supervised training and analysis by many common metrics that rely on pixel-to-pixel comparisons against groundtruths.

\texttt{PaperFactory} was used with 300 textures, dramatically increasing the possible variations in the final dataset.

\emph{Augraphy} by design includes a high degree of randomness: most augmentations make several calls to a pseudorandom number generator, on which several transforms depend.
All or none of these augmentations may have applied to each input image, and only one pipeline run was completed per image.

The conjunction of all these factors implies \emph{ShabbyPages} is a sample from a massive space.
Indeed, the cosine similarity computed over OpenAI's CLIP representations of each output image averages to 0.49, much higher than \emph{NoisyOffice}'s 0.31.

\begin{figure}
    \includegraphics[width=1.0\textwidth]{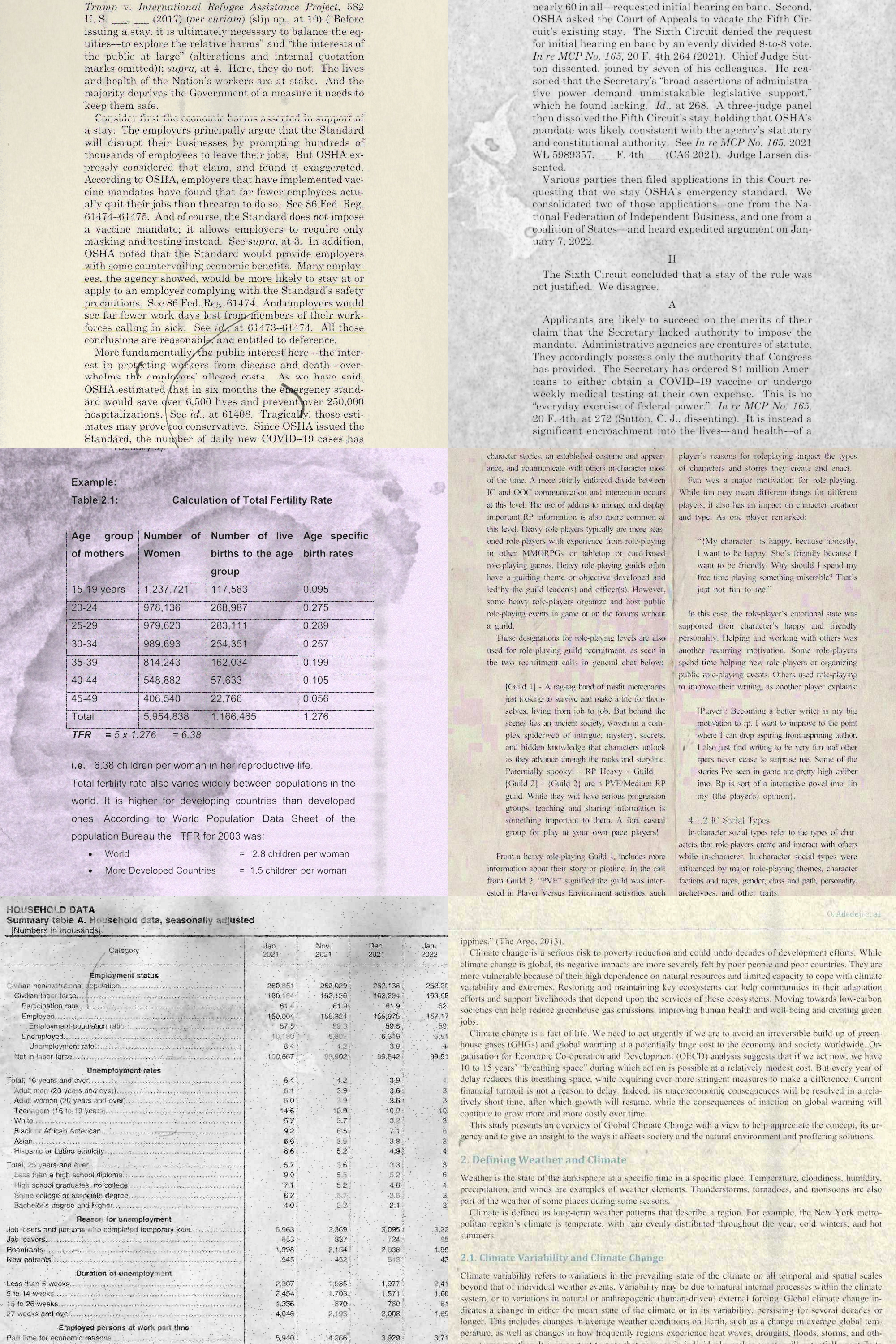}
    \caption{Sample noisy patches from \emph{ShabbyPages}.}
    \label{fig:matrix_tall2}
\end{figure}

\subsubsection{Processing \emph{Augraphy} Pipelines.}
Execution time is dependent on which augmentations are executed at runtime; a long \emph{Augraphy} pipeline can take several seconds to process large images.
The library is under active development with performance enhancements underway, but generating large datasets sequentially can be unnecessarily time consuming when dealing with thousands of files, so we use a multi-process pooling technique to distribute the workload across many processor execution threads.
For each process, a new pipeline with new probability coefficients was generated, representing a nontrivially-different sequence of document operations.
Each of the 6202 base images was passed through a unique pipeline, with the output saved to a separate directory.
\emph{Augraphy}'s API design made it very easy to write the code to accomplish this; the entire \emph{ShabbyPages} generation script is fewer than 30 lines long including imports, with the most critical piece represented below:

\begin{lstlisting}[language=Python]
  def run_pipeline(filename):
    image = cv2.imread(filename)
    # returns the current ShabbyPages pipeline
    pipeline = get_pipeline()
    data = pipeline.augment(image)
    shabby_image = data['output']
    cv2.imwrite(filename + '-shabby.png',
                shabby_image)
\end{lstlisting}

Processing all 6202 images took less than half an hour on the 64 cores of a Graviton3 c7g.16xlarge instance on AWS. In testing, we rendered several tens of thousands of such images, with the cumulative computation time still less than one day due to \emph{Augraphy}'s efficiency.

\subsection{\emph{ShabbyReal}}
In keeping with the spirit of \emph{NoisyOffice}, our original inspiration, and to provide out-of-distribution test data for ours and other projects, we developed a separate but related dataset which we call \emph{ShabbyReal} that contains 200 real-world clean/dirty document image pairs. To build this, we isolated reproduction steps for several physical operations, designed real-world augmentation pipelines with these, and then engaged a team of crowdworkers to use these instructions to produce several copies of printed born-digital documents with real noise.

Five individuals were recruited for this task, which involved printing 40 pages of documents and manually applying various effects such as folds, wrinkles, smudges, highlighters, pencils, crayons, splatters of ink or other liquids, printing, scanning, faxing and other creative and reasonable treatments determined by these workers.  Post processing was also required to align scanned dirty images to their original groundtruths to make them suitable for denoising or binarization tasks, effectively doubling the product cost of the dataset. 
The process continued over a period of two weeks, incurring a cost of approximately \$2,000 USD.
While there are plenty of opportunities to streamline the operation, scaling this process is expensive in terms of both time and money, and may have unpredictable results.

Several challenges arose in the construction of this dataset.
Old fax machines, faulty commercial printers, and other hardware suitable for inducing the intended effects were difficult to source, and frequently non-functional when they could be found.
Environmental concerns like the volume of paper and ink used, and the energy expenditure both in electricity and human terms, also factored into our decision to pursue synthetic augmentations for \emph{ShabbyPages}.
Manually producing a database of noisy documents is a time-consuming and repetitive task; creating just 200 real-world documents with real noise features required the labor of several people.
Nevertheless, despite the limited size of the resulting dataset, the scarcity of real-world datasets with ground truth images makes the \emph{ShabbyReal} corpus extremely valuable; its inclusion in \emph{ShabbyPages} even more so.
This type of dataset is critical for benchmark comparison with synthetically-augmented document images, and for other tasks such as denoising and binarizing: we use \emph{ShabbyReal} later for a human-perception evaluation of denoising performance.

\begin{figure}
\centering
\includegraphics[width=0.98\columnwidth]{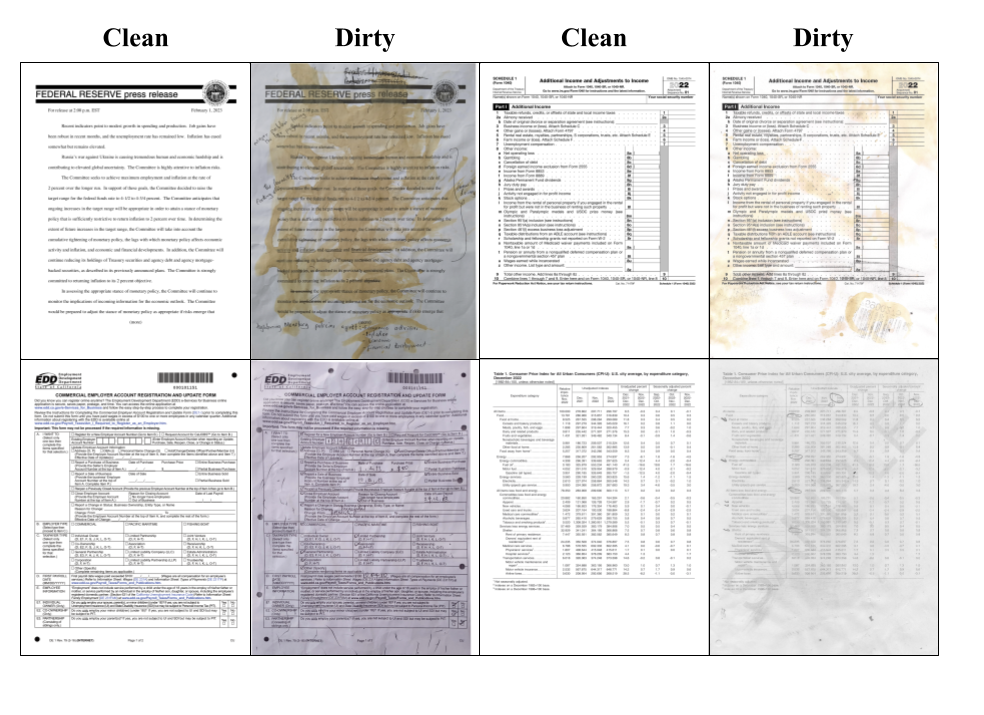}
\caption{Example \emph{ShabbyReal} images and their clean counterparts.}
\label{fig:shabbyreal_sample}
\end{figure}

\section{Experimentation with the \emph{ShabbyPages} Set}
Denoising and binarization are two important techniques for the removal of unwanted data from images.
Both approaches can be achieved by supervised learning of relationships between features of the input and output data.
In this section, we train two NAFNet \cite{ref_nafnet} instances as denoisers on both the \mbox{\emph{NoisyOffice}} and the \mbox{\emph{ShabbyPages}} datasets. We evaluate these models' predictions in a cross-validation test on both datasets, both by visual inspection and by metric computation.
We then perform a similar experiment where the same NAFNets are trained to not only remove noise from the test images, but also to classify the foreground and background pixels, predicting a binarized clean document.
The model was ``off-the-shelf''; besides rigging up the training inputs and choosing a reasonable training epoch limit, we made no significant alterations to the provided hyperparameters.

\subsection{Denoising}
Digital computation has enabled humanity to produce ever-growing amounts of both data and reasons to process it.
Doing so is easiest when the data is free of unexpected outliers, the signal has less jitter, and perhaps most importantly, when our aesthetic sensibilities are unchallenged.

Denoising grayscale images is a challenging task, as models must determine not only which features should appear in the output, but also the 8-bit pixel intensity over the whole image.
We randomly selected ten patches of 400x400 pixels from each page, then trained another NAFNet instance on this data for just 14 epochs (about 22 hours of compute time).
For this task, we considered the structural similarity index (SSIM), a common image processing metric which takes into account visually-perceptible data such as luminance and contrast, appropriate for grayscale images.
The results of the denoising cross-validation experiment are presented in the table below.

\begin{table}[]
    \centering
    \caption{Document image denoising cross-test performance of NAFNet models trained on \emph{ShabbyPages} and \emph{NoisyOffice}.} \label{tab:denoising_results}
    \scalebox{0.9}{
    \begin{tabular}{llll}
    \toprule
    \textbf{Training Set~~} & \textbf{Test Set} & \textbf{Model} & \textbf{SSIM$\uparrow$~}\\
        \midrule
        \textsf{\emph{ShabbyPages}} & \textsf{\emph{NoisyOffice-real}} & NAFNet & \textbf{0.96}\\
        \textsf{\emph{NoisyOffice-sim}} & \textsf{\emph{ShabbyPages}} & NAFNet & 0.83\\
        \bottomrule
    \end{tabular}}
\end{table}

As can be seen from Table~\ref{tab:denoising_results}, the \emph{ShabbyPages}-trained denoiser generalized to \emph{NoisyOffice} far better than the \emph{NoisyOffice}-trained model generalized to \emph{ShabbyPages}. This result is consistent with the much higher diversity present within the \emph{ShabbyPages} set, relative to features found in the \emph{NoisyOffice} database.

Machine learning is ultimately about getting computers to complete tasks which normally require human labor; as further validation for the efficacy of \emph{ShabbyPages} in training performant denoisers, we also used our \emph{ShabbyPages}-trained NAFNet model to denoise the \emph{ShabbyReal} set, and present some examples of its performance below.

\begin{figure}
\centering
\includegraphics[width=0.98\columnwidth]{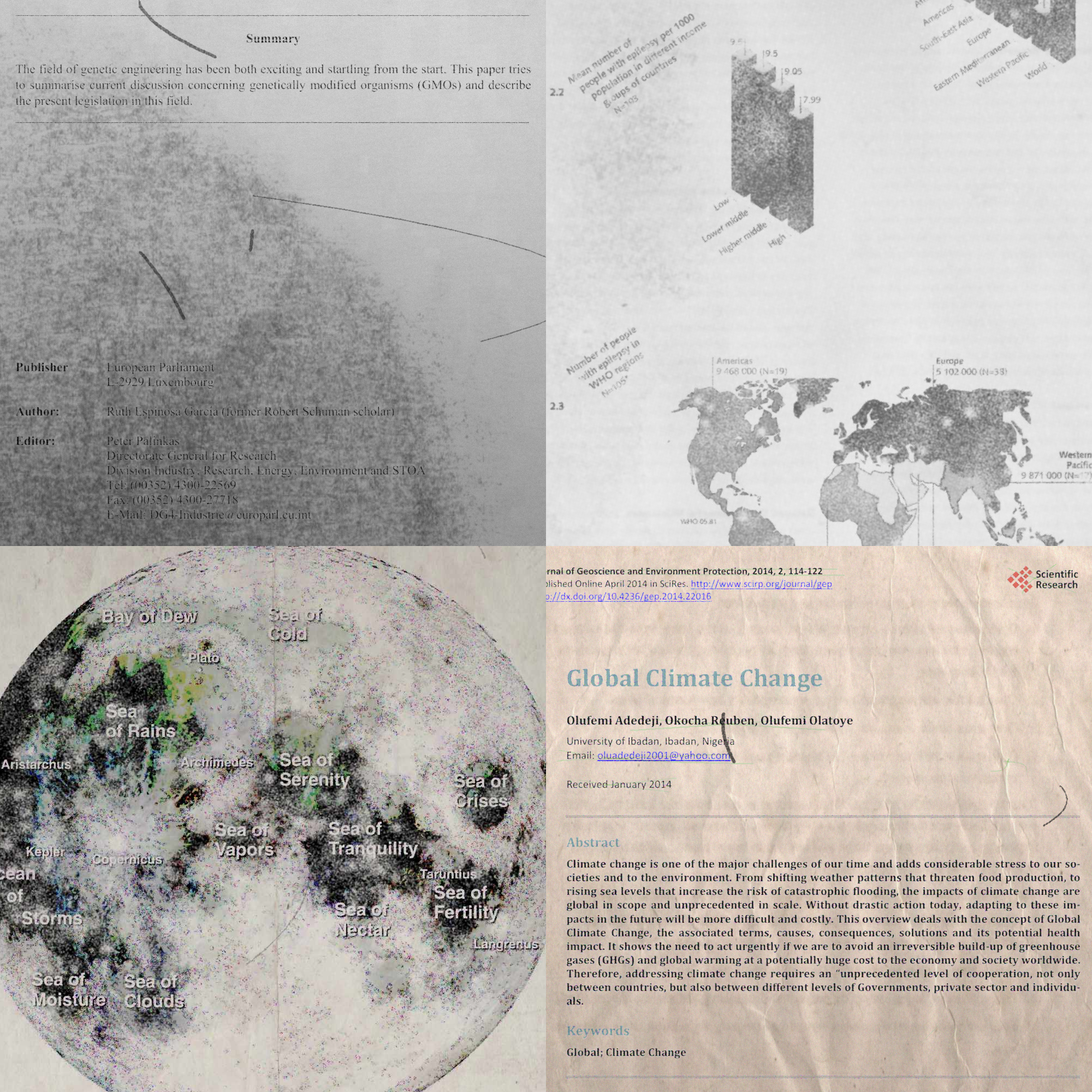}
\caption{Sample patches from \emph{ShabbyPages} dataset.}
\label{fig:shabbyreal_denoising}
\end{figure}

\subsection{Binarization}
In this section, we discuss results obtained by training a NAFNet denoiser to both restore augmented images to their ground truths and to classify pixels as foreground.

We first performed an ensemble binarization preprocessing step on the clean groundtruth images.
For each image, we computed the Niblack, Sauvola, and Otsu thresholds, using these to binarize three copies of the input. The average over the resulting matrices was taken elementwise, producing the final binarized groundtruth.

One instance of the model was trained on the full \emph{SimulatedNoisyOffice} corpus, while the other was trained on a 1050-image subset of \emph{ShabbyPages}, cropped to a single random 500x500 patch. In both cases, the prediction target was the relevant ensemble-binarized groundtruths, and each instance trained for 100 epochs.
As a test, these models were used to denoise and binarize both the full \emph{NoisyOfficeReal} dataset and a 450-image testing subset taken from \emph{ShabbyPages} which does not overlap the training subset.

Multiple common image processing metrics were applied to the predictions made by each model, comparing these to the preprocessed groundtruths.
In addition to the structural similarity metric from the denoising task, we add the peak signal to noise ratio (PSNR) and the root mean squared error (RMSE). Both of these metrics are more useful for determining the differences between binary images: together, they give some idea of the degree to which noise artifacts remain in the image after denoising and binarization affect the visual signal, and how many such artifacts exist.
The averages of each metric over all predictions was taken, with the results presented in Table~\ref{tab:binarization_results}.

\begin{table}[]
    \centering
    \caption{Document image binarization performance of a NAFNet model trained and tested on \emph{ShabbyPages} and \emph{NoisyOffice}.}
    \scalebox{0.9}{
    \begin{tabular}{llllll}
    \toprule
    \textbf{Training Set~~} & \textbf{Test Set} & \textbf{Model} & \textbf{SSIM$\uparrow$~} & \textbf{PSNR$\uparrow$~} & \textbf{RMSE$\downarrow$~}\\
        \midrule
        \textsf{\emph{ShabbyPages}} & \textsf{\emph{NoisyOffice-real}} & NAFNet & \textbf{0.947} & \textbf{38.098} & \textbf{3.205}\\
        \textsf{\emph{NoisyOffice-sim}} & \textsf{\emph{ShabbyPages}} & NAFNet & 0.811 & 34.562 & 5.384\\
        \bottomrule
    \end{tabular}}
    \label{tab:binarization_results}
\end{table}

We cross-validate these models by predicting cleaned and binarized images using the other testing set as inputs.
The \emph{ShabbyPages}-trained model was able to classify foreground and background pixels in the \emph{NoisyOffice} test set with a substantially higher degree of accuracy than the \emph{NoisyOffice}-trained model could classify \emph{ShabbyPages}, even with less-favorable training data.
\emph{ShabbyPages} contains a much higher feature diversity than \emph{NoisyOffice}, so these results were unsurprising, although we did expect our model to achieve worse results than it did, since it only saw a single patch from each full page.
The \emph{ShabbyPages}-trained model performs well on a much wider range of features, handling tables, graphs, and even simple images quite well.
This is a strong indicator that neural networks trained on \emph{ShabbyPages} can outperform those trained on \emph{NoisyOffice} when generalizing to other datasets.

\section{Conclusion}
Supervised learning requires a collection of training data and accompanying ``labels'', which in the graphical modality are clean images, free of noise or other degradations.
We presented \emph{ShabbyPages}, a large dataset composed of 6202 clean/noisy pairs of synthetically-noisy images taken from 600 digital documents.
Information about the set's contents and its relation to other common datasets was explored.
The noisy images were created from their clean origins by the application of a pipeline developed with the \emph{Augraphy} library; the production of \emph{ShabbyPages} was detailed in the third section, with all source code and input materials made openly available.
\emph{ShabbyPages} was used to train performant denoising and binarization models which generalize well to document images containing real noise.

\bibliographystyle{splncs04}
\bibliography{paper}
\end{document}